\pdfoutput=1

\documentclass[11pt]{article}

\usepackage{acl}
\usepackage{times}
\usepackage{latexsym}
\usepackage{graphicx}
\usepackage{amsmath}
\usepackage{adjustbox}
\usepackage{makecell}
\usepackage{booktabs}
\usepackage[T1]{fontenc}
\usepackage[utf8]{inputenc}
\usepackage{microtype}

\title{Why only Micro-$F_1$? Class Weighting of Measures for Relation Classification}

\author{David Harbecke$^\clubsuit$, Yuxuan Chen$^\clubsuit$, Leonhard Hennig$^\clubsuit$, Christoph Alt$^{\spadesuit\heartsuit}$ \\
  $^\clubsuit$German Research Center for Artificial Intelligence (DFKI), Berlin \\
  $^\spadesuit$Humboldt Universität zu Berlin
  ~~~$^\heartsuit$Science of Intelligence\\
  $^\clubsuit$\texttt{\{firstname\}.\{lastname\}@dfki.de} \\
  $^\spadesuit$\texttt{christoph.alt@posteo.de}
}

\begin{document}
\renewcommand{\arraystretch}{1.1}
\renewcommand{\cellalign}{tl}
\hyphenation{Sem-Eval}

\newcommand{\insertdatasetsinfotable}{
    \begin{table*}[htb]
    \setlength{\tabcolsep}{12pt}
        \centering
            \begin{tabular}{@{}lrrrr@{\hskip 5pt}rrl@{}}
             \toprule
             & & & & \multicolumn{2}{l}{Perplexity} & & \\
             Dataset & \#Rel & \#Samples & \%NA & w NA & w/o NA & Ratio & Evaluation \\
             \midrule
             TACRED & 42 & 106264 & 79.5 & 3.31 & 23.39 & 250 & micro-$F_1$ \\
             NYT & \makecell[r]{53 \\ 24} & \makecell[r]{694491 \\ 66194 } & \makecell[r]{79.4 \\ 0} & \makecell[r]{1.27 \\ 6.24 } & \makecell[r]{7.84 \\ 6.24} & \makecell[r]{ 2793 \\ 2485 } & precision at $k$, AUC\\
             ChemProt & 13 & 10065 & 0 & 7.23 & 7.23 & 314 & micro-$F_1$  \\
             DocRED & 96 & 50503 & 0 & 33.13 & 33.13 & 2837 & micro-$F_1$, AUC \\
             SemEval & \makecell[r]{19 \\ 10} & \makecell[r]{10717 \\ 10717} & \makecell[r]{17.4 \\ 17.4} & \makecell[r]{14.45 \\ 9.61} & \makecell[r]{14.37 \\ 8.80} & \makecell[r]{291 \\ 2.10} & \makecell{macro-$F_1$ (official),\\ micro-$F_1$ (popular)}\\
             \bottomrule
            \end{tabular}
        \caption{Statistics for popular RC datasets.
        The number of relations, samples and percent of negative samples are for the whole dataset.
        Perplexity of the classes is given for the test set, with and without negative samples.
        This value would be equal to \#Rel for a fully balanced dataset.
        Ratio is between the counts of the most and least frequent positive class of the test set.
        We also list the popular evaluation methods.
        The upper line for NYT indicates the original dataset by \citet{riedel2010modeling}, the lower line is the frequently used version by \citet{hoffmann2011knowledge}.
        The upper SemEval entry considers the direction between the nominals, the lower one does not.}
        \label{tab:datasets_info_table}
    \end{table*}
}

\newcommand{\insertweightingexplaintable}{
    \begin{table*}[htb]
    \setlength{\tabcolsep}{10pt}
        \centering
            \begin{tabular}{@{}lll@{}}
             \toprule
             Method & Formula & Focus \\
             \midrule
             Micro & - & calculation over dataset, class membership is not considered \\
             Weighted & $n_i$ & weighting all classes by instance count, similar to micro \\
             \textbf{Dodrans} & ${n_i}^{3/4}$ & evaluating closer to generalization performance  \\
             \textbf{Entropy} & $-n_i \cdot \log_2(n_i/\sum_j n_j)$ & reducing impact of data distribution on evaluation \\
             Macro & $1$ & equal weighting of all classes \\
             \bottomrule
            \end{tabular}
        \caption{Weighting schemes for evaluation of multi-class classification.
        $n_i$ indicates the count of elements for class $i$ and the Formula column shows the weight the class is assigned before normalization. 
        The metrics are loosely ordered from top to bottom with the higher entries focusing more on instances and the lower entries focusing more on class membership.
        This usually corresponds to the model score, it is rare that models are better on classes with fewer samples.
        Methods in bold are proposed by us.}
        \label{tab:weighting_table}
    \end{table*}
}

\newcommand{\inserttacredweightsfigure}{
 \begin{figure}[tb]
  \centering
  \includegraphics[width=\columnwidth]{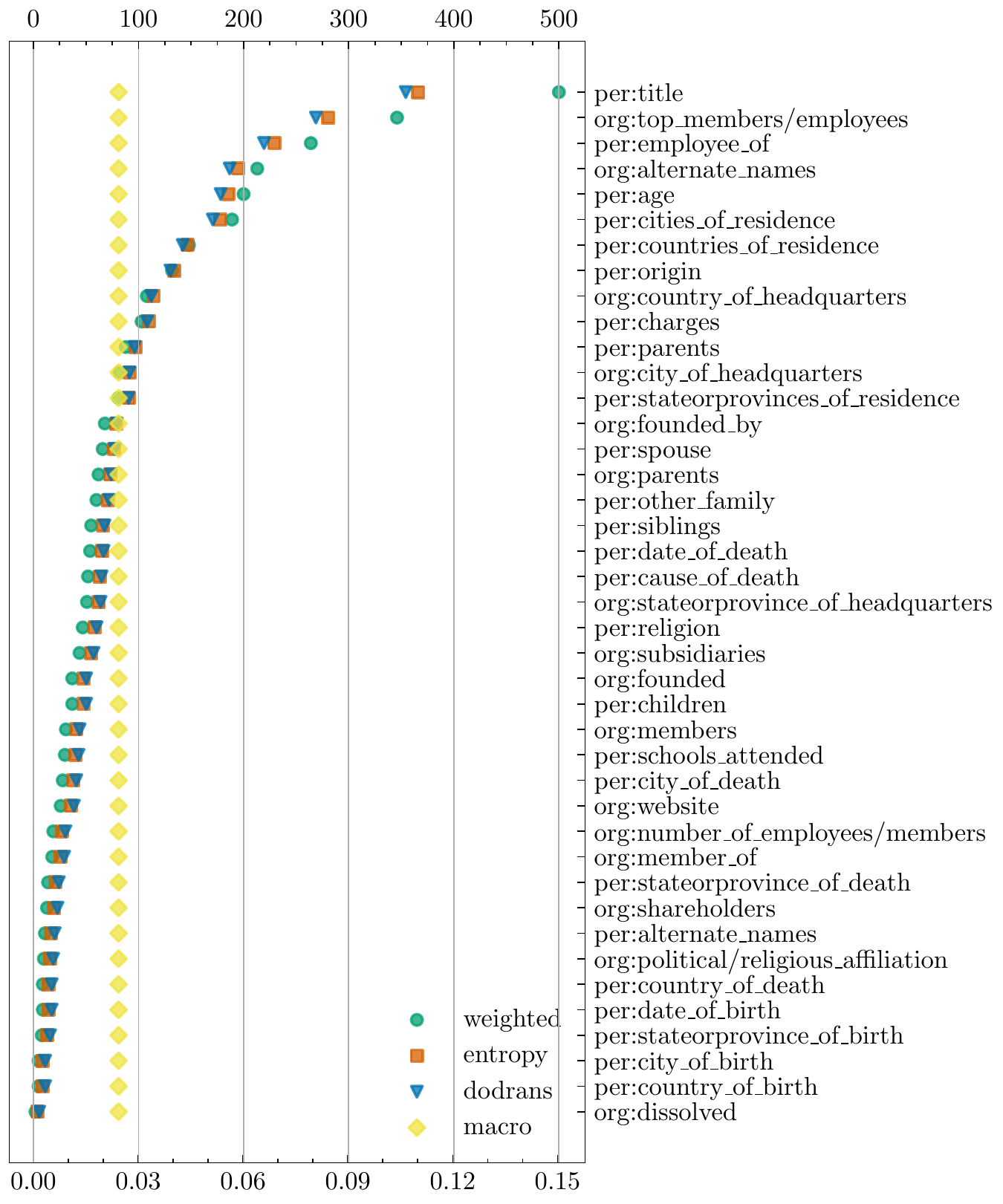}
  \caption{TACRED relations and their respective weights under different weighting schemes.
  The lower x-axis denotes the normalized weight given to a relation for a scheme.
  The upper x-axis corresponds to the counts of the relations in the test set for the class-weighted scheme.
  The y-axis denotes all positive relations.
  The negative \emph{NA} class is not listed and has 12184 samples.
  The entropy and dodrans weighting scheme produce similar weights and are between weighted and macro weighting.}
  \label{fig:tacred_weights}
 \end{figure}
}

\newcommand{\inserttacredresultstable}{
    \begin{table*}[htb]
        \setlength{\tabcolsep}{10pt}
        \centering
            \begin{tabular}{@{}llllll@{}}
             \toprule
             Method & {Micro} & {Weighted} & {Dodrans} & {Entropy} & {Macro} \\
             \midrule
             RECENT & 71.5\small{$\pm$0.4}  & 67.8\small{$\pm$0.4} & 62.5\small{$\pm$0.4} & 63.6\small{$\pm$0.4} & 43.1\small{$\pm$0.6} \\
             PTR & 72.5\small{$\pm$0.3} & 72.1\small{$\pm$0.5} & 69.8\small{$\pm$0.5} & 70.3\small{$\pm$0.5} & 60.3\small{$\pm$0.8} \\
             \midrule
             $p$-value & $3\cdot10^{-3}$ & $3\cdot10^{-6}$ & $10^{-8}$ & $2\cdot10^{-8}$ & $2\cdot10^{-10}$ \\
             Cohen's $d$ & $2.8$ & $8.7$ & $14.8$ & $13.5$ & $24.2$ \\
             \bottomrule
            \end{tabular}
        \caption{TACRED $F_1$-scores with different weighting schemes.
        Positive scores indicate PTR performs better than RECENT for all weighting schemes.
        The difference is smallest for the micro and largest for the macro weighting scheme.
        All $p$-values are smaller than $\alpha=0.05$.
        All effect sizes are huge ($>2.0$) under \citet{sawilowsky2009new}'s rules of thumb.}
        \label{tab:tacred_results_table}
    \end{table*}
}

\newcommand{\insertsemevalresultstable}{
    \begin{table*}[htb]
        \setlength{\tabcolsep}{10pt}
        \centering
            \begin{tabular}{@{}llllll@{}}
             \toprule
             Method & {Micro} & {Weighted} & {Dodrans} & {Entropy} & {Macro} \\
             \midrule
             BERT\textsubscript{EM} & 89.1\small{$\pm$0.3} & 89.1\small{$\pm$0.3} & 88.7\small{$\pm$0.3} & 88.6\small{$\pm$0.3} & 82.7\small{$\pm$0.4}\\
             PTR & 88.4\small{$\pm$0.3} & 88.3\small{$\pm$0.3} & 88.1\small{$\pm$0.3} & 88.0\small{$\pm$0.3} & 87.8\small{$\pm$0.5} \\
             \midrule
             $p$-value & $0.005$ & $0.006$ & $0.023$ & $0.023$ & $7\cdot10^{-8}$ \\
             Cohen's $d$ & -2.5 & -2.4 & -1.8 & -1.8 & 11.5 \\
             \bottomrule
            \end{tabular}
        \caption{SemEval $F_1$-scores with different weighting schemes.
        The directionality is of the relations is considered, s.t. there are 19 classes, the negative class is not included in evaluation.
        Negative scores indicate BERT\textsubscript{EM} performs better, positive scores indicate PTR performs better.
        All $p$-values are smaller than $\alpha=0.05$.
        All absolute effect sizes are very large ($>1.2$) or huge ($>2.0$).}
        \label{tab:semeval_results_table}
    \end{table*}
}

\maketitle

\begin{abstract}
Relation classification models are conventionally evaluated using only a single measure, e.g., micro-$F_1$, macro-$F_1$ or AUC.
In this work, we analyze weighting schemes, such as \emph{micro} and \emph{macro}, for imbalanced datasets.
We introduce a framework for weighting schemes, where existing schemes are extremes, and two new intermediate schemes.
We show that reporting results of different weighting schemes better highlights strengths and weaknesses of a model.
\end{abstract}

\section{Introduction}
Relation classification (RC) models are typically compared with either micro-$F_1$ or macro-$F_1$, often without discussing the measure's properties (see e.g.\ ~\citealp{zhang2017position, yao2019docred}).
Each measure highlights different aspects of model performance~\cite{sun2009classification}.
However, using an inappropriate measure can lead to the preference of an unsuitable model~\cite{branco2016survey}, e.g., tasks with an imbalanced or long-tailed class distribution.
We argue that model evaluation should better reflect this, particularly as rare phenomena become more important in NLP~\cite{rogers2021changing}.

\insertdatasetsinfotable

For instance, popular datasets for RC, such as TACRED~\cite{zhang2017position}, NYT~\cite{riedel2010modeling}, ChemProt~\cite{kringelum2016chemprot}, DocRED~\cite{yao2019docred}, and SemEval-2010 Task 8~\cite{hendrickx2010semeval}, often exhibit a highly imbalanced label distribution (see Table~\ref{tab:datasets_info_table} and, e.g., the TACRED class distribution\footnote{ \url{https://nlp.stanford.edu/projects/tacred/\#stats}}).
The main reasons are the natural data imbalance, i.e.\ the occurrence frequency of relation mentions in text, as well as the incompleteness of knowledge graphs like Freebase~\cite{bollacker2008freebase} used in distantly supervised RC.
For example, 58\% of the relations in the NYT dataset~\cite{riedel2010modeling} have fewer than 100 training instances~\cite{han2018fewrel}, and the most frequent relation \emph{location/contains} is assigned to 48.3\% of the positive test instances.
However, for applying RC to real-world problems, it is especially important to discover instances of relations that are not yet covered well in a given knowledge base.

Table \ref{tab:datasets_info_table} lists statistics of the aforementioned RC datasets, including their perplexity and common evaluation measures.
TACRED and the original version of NYT contain predominantly negative samples\footnote{Negative samples in RC means none of the dataset's relations hold.
Depending on the dataset, this class is coined \emph{no-relation}, \emph{NA} or \emph{Other}.
We use negative class or \emph{NA}.}.
All datasets, except for undirectional SemEval, exhibit a large ratio between most frequent and least frequent positive class in the test set.
The perplexity of test set distributions is also much lower than the relation count for all datasets except SemEval.
Reporting only a single measure therefore cannot exhaustively capture model performance on these datasets, especially for the long tail of relation types.
For example,~\citet{alt-etal-2019-fine} show that on the NYT dataset, AUC scores and P-R-Curves of several state-of-the-art models are heavily skewed towards the two most frequent relation types \emph{location/contains} and \emph{person/nationality}. 
TACRED, ChemProt, DocRED and SemEval results are usually only reported in micro-$F_1$, which does not consider class membership.

In this paper, we introduce a framework for weighting schemes of measures to address these evaluation deficits.
We present and motivate two new weighting schemes that are in between the extremes of micro- and macro-weighting.
We demonstrate these, micro-, class-weighted- and macro-$F_1$ on TACRED and SemEval with two popular models each.
We show that more information about models can be inferred from our results and point out what further steps should be taken to improve evaluation in relation classification.

\section{Methods}
We first give background on the $F_1$-score and existing $F_1$ weighting schemes.
We present our framework of weighting schemes.
We introduce two new weighting schemes.
Finally, we outline statistical tests.

\subsection{Background}
The $F_{\beta}$-score~\cite{rijsbergen1979information, lewis1994sequential} calculates a score in the interval $[0, 1]$ through the formula
\begin{equation}
    F_{\beta} = \frac{(1 + \beta^2) \cdot TP}{(1 + \beta^2) \cdot TP + \beta^2 \cdot FN + FP}
    \label{eq:F1}
\end{equation}
with the true positives (TP), false negatives (FN) and false positives (FP) of a confusion matrix.
This definition is identical to the weighted harmonic mean of precision and recall.
The positive coefficient $\beta$ is used as a trade-off between the error types FN and FP.
If there is no preference known or pre-determined, this coefficient is usually set to 1.
In multi-class classification the confusion matrix can either be calculated once for the whole dataset, or separately for each class.
The former method yields micro-$F_1$.

\insertweightingexplaintable
\textbf{Micro} weighting does not consider class membership for any test sample.
If the predictions and labels of all classes are considered, micro-$F_1$ is equal to accuracy, as the denominator in Eq.\ \ref{eq:F1} is twice the dataset.
In RC, the TP of the negative class are usually not considered, in which case micro-$F_1$ is not equal to accuracy.
For the $F$-score, \emph{micro} is the only weighting where the impact of a sample on the score is not conditioned on the model performance on the rest of the class~\cite{forman2010apples}.
If the test set is considered to have a representative data distribution, the micro-weighted score is a frequentist evaluation of model performance.

There exist two other ways to calculate and combine $F_1$-scores for a multi-class problem.
First, multi-class $F_1$-scores can be calculated for each class and then a weighted average class score is taken.
Second, precision and recall scores for each class can be calculated and weighted, then the harmonic mean of weighted precision and weighted recall is taken.
\citet{opitz2019macro} show that the first method is more robust and less favorable to biased classifiers.
We use this method in our proposed framework.

\textbf{(Class-)weighted}-$F_1$ is similar to micro-$F_1$.
$F_1$-scores are calculated for each class individually and then weighted by the class count.
Thus, both schemes approximately weigh all samples equally.

\textbf{Macro} weighting gives an equal weight for each class with positive sample count regardless of the specific sample count.
This gives information about model performance if class imbalance is not considered.

In general, there is a correspondence between training loss and evaluation measure~\cite{li2020dice}.
One disadvantage of multiple weighting schemes is that each weighting scheme can be optimized for.
To achieve a better score for a specific weighting, class weights could be set proportional to the weighting of the class during training.
However, we argue that model results should always be presented with multiple weightings for one dataset.
Especially, when comparing different models all weightings should be reported for each model.
This can clarify whether a model is good for all weightings or just \emph{micro} or \emph{macro}.
Furthermore, with datasets that are currently evaluated with different weightings, it is easier to identify whether a model is specifically good for a dataset or for a weighting.

\subsection{Framework for Weighting Schemes}
We discuss a framework that summarizes the rules we give to class-weighting schemes.
Then we introduce two new class weighting schemes.
All discussed weighting schemes can be found in Table~\ref{tab:weighting_table}.
They are independent of the measure that is used to calculate a score for each class.

\emph{(Class-)weighted} and \emph{macro} weighting are the extremes of ``degressive proportionality''\footnote{\url{https://eur-lex.europa.eu/legal-content/EN/TXT/HTML/?uri=CELEX:32013D0312&from=EN\#d1e114-57-1}} or ``allocation functions''~\cite{slomczynski2012mathematical}.
These are, e.g., used by the European Parliament to allocate seats to member nations depending on the population of the nation.
They state that allocation should be monotonic increasing (see \ref{eq:D1}) and proportionally decreasing (see \ref{eq:D2}).
To adopt this to a weighting scheme for multi-class evaluation, we add a normalizing desideratum that determines the sum of weights over all classes to be 1 (see \ref{eq:D0}).

Let $n_i>0$ be the count of samples of class $i$ and $w_i \geq 0$ the weight assigned to the score of class $i$.
We have the following desiderata:
\begin{subequations}
\nonumber
\begin{align*}
  \tag{D0} &\sum_i w_i = 1  \label{eq:D0} \\
  \tag{D1} & n_i \geq n_j \Rightarrow w_i \geq w_j \label{eq:D1}\\
  \tag{D2} & n_i \geq n_j \Rightarrow \frac{w_i}{n_i} \leq \frac{w_j}{n_j} \label{eq:D2}
\end{align*}
\end{subequations}

Note that these desiderata do not restrict the scoring function that assigns scores $s_i$ to class $i$.
The weighted evaluation score is then given by $\sum_i w_i s_i$.

\subsection{Weighting Schemes}

\textbf{Macro}: Macro weighting is one extreme by setting equality on the weights of desideratum \ref{eq:D1}.
It implies that we do not consider the instance counts per class, but treat all classes equally.

\noindent \textbf{(Class-)weighted}: Class-weighted is the other extreme by setting equality on the fraction of weights and counts in desideratum \ref{eq:D2}.
It implies that we do not consider class constituency but weight all samples equally.

\noindent \textbf{Dodrans}:~\citet{cao2019learning} demonstrate that their balanced generalization error bound for binary classifiers in the separable case can be optimized by setting margins proportional to ${n_i}^{-1/4}$.
They use this derivation from a limited theoretical scenario to improve the performance of several classifiers on imbalanced multi-class datasets.
A term proportional to ${n_i}^{-1/4}$ is added in the loss function.
While this added term is not directly transferable, we propose adapting this as a multiplicative factor in weighting classes for multi-class evaluation:
$w_i \propto {n_i}^{-1/4}n_i = n_i^{3/4}.$
We coin this weighting \emph{dodrans} (``three-quarter'').

\noindent \textbf{Entropy}: We also want to provide a weighting scheme that takes into consideration how hard a class is to predict.
To this end, we propose weighting classes proportional to their term in the Shannon entropy formula
\setcounter{equation}{1}
\begin{align}
    H(X) &= - \sum_{i} P(x_i) \log(P(x_i))\\
    w_i &\propto P(x_i) \log(P(x_i)).
\end{align}
We interpret $P(x_i)$ for class $i$ to be the probability of it appearing in the dataset, s.t. $P(x_i)=n_i/\sum_j n_j$.
Thus, without normalization the model score is now the sum over all classes of the model performance on a class times the difficulty and frequency of the class.
Note, that this weighting scheme does not fulfil desideratum \ref{eq:D1}, since it is decreasing for classes $i$ with $P(x_i) > e^{-1}$.
This is related to the fact that classes that are too large become easier to predict for a model, the model can just default to predicting this class.
It can also be desirable that a class does not gain relative importance once it contains more than half of the dataset.
For RC, this often has little consequence.
If we include \emph{NA} in the normalization, it is usually the largest class and other classes are below an $e$-th of the dataset.
Table \ref{tab:weighting_table} shows an overview of the mentioned schemes.

Figure \ref{fig:tacred_weights} displays the weights that these schemes assign to the classes of the TACRED test set.
The \emph{weighted} scheme is proportional to class counts and produces the most imbalanced weights.
\emph{Dodrans} and \emph{entropy} produce slightly more balanced weights and differ from \emph{weighted} for the most frequent classes.
\emph{Macro} considers all classes equally, regardless of class count.

\subsection{Statistical Testing}
Currently, most RC works report a single score for each dataset.
This can be the result from a single run or the median score from multiple runs.
However, this does not allow to measure how large the difference between models is.
Recently, analysis papers in NLP have recorded mean and standard deviation over multiple runs~\cite{madhyastha2019model, zhou2020curse}, as this allows for statistical tests.

\inserttacredweightsfigure

We first test for significance and report $p$-values.
We employ Welch's $t$-test to test the hypothesis that the models have equal mean. 
Following~\citet{zhu2020nlpstattest}, we also report Cohen's $d$ effect size to determine how large the difference between models is for a specific measure.
For two models with the same number $n>1$ of runs, Cohen's $d$ is given by
\begin{equation}
    d = \sqrt{2}\frac{\mu_1 - \mu_2}{\sqrt{\sigma_1^2 + \sigma_2^2}}
\end{equation}
with $\mu_i$ and $\sigma_i^2$ being mean and variance of model $i$'s scores.
We do this, as two different models never perform exactly the same, i.e.\ significance just depends on the number of runs and we also want to score the difference between the models.

\inserttacredresultstable
\insertsemevalresultstable

\section{Experiments}
We evaluate and compare three RC methods with our proposed measures on two datasets.
We choose these methods, as RECENT~\cite{lyu2021relation} and BERT\textsubscript{EM}~\cite{soares2019mtb} are based on vanilla fine-tuning of a pre-trained language model, with a classification head on top.
PTR~\cite{han2021ptr} is based on prompt-tuning.
RECENT and PTR report similar micro-$F_1$ performance on TACRED, as do BERT\textsubscript{EM} and PTR on SemEval.
In this way we can compare performance of the two paradigms for other weightings.

RECENT proposes a model-agnostic paradigm that exploits entity types to narrow down the candidate relations.
Given an entity-type combination, a separate classifier is trained on the restricted classes.
\citet{soares2019mtb} compare various strategies that extract relation representation from Transformers and claim \textsc{Entity Start} (i.e.\ insert entity markers at the start of two entity mentions) yields the best performance.
PTR also takes entity types into consideration and constructs prompts composed of three subprompts, two corresponding to the fill-in of the entity types and one predicting the relation.

In our experiments we use RECENT\textsubscript{GCN} for RECENT, BERT\textsubscript{EM} with \textsc{Entity Start}, and unreversed prompts for PTR.
We use the official repositories for RECENT and PTR, we reimplement BERT\textsubscript{EM}\footnote{Our reimplementation is available at \url{https://github.com/dfki-nlp/mtb-bert-em}.}.
We use the hyperparameters proposed in the original papers and conduct five runs for each model.
Additional implementation and training details can be found in Appendices \ref{sec:appendix-implementation} and \ref{sec:appendix-training}.

The main focus is unearthing performance information about these methods that was previously obscured by single score measures.
The number of weighting schemes does not influence the computational cost, as each score is determined through the predictions in a run and does not require specific tuning.\footnote{We provide a package to add these scores to a Scikit-learn~\cite{pedregosa2011scikit} classification report at \url{https://github.com/DFKI-NLP/weighting-schemes-report}.}
We acknowledge that each weighting scheme could be optimized for during training which gives additional importance to reporting multiple measures for each model.

\subsection{Results}
Table \ref{tab:tacred_results_table} shows results for TACRED.
PTR significantly outperforms RECENT across all weighting schemes.
The difference between the models is smallest for micro-$F_1$ and increases for all schemes that weigh classes more equally.
For macro-$F_1$ the difference is starkest with effect size $24.2$.

Table \ref{tab:semeval_results_table} displays results for SemEval.
BERT\textsubscript{EM} significantly outperforms PTR in the micro-$F_1$ measure and all other weightings except for macro-$F_1$.
All effect sizes are either large or huge, by far the largest effect size is between PTR and BERT\textsubscript{EM} regarding macro-$F_1$ though.
The SemEval test set contains a single sample of the \emph{Entity-Destination(e2,e1)} class which is quite impactful for the macro-$F_1$ of the models but has negligible impact on all other weighting schemes.
The scores from \emph{dodrans} and \emph{entropy} indicate that only if all classes are considered equally important the PTR model should be preferred.
This indicates that either the PTR model learns almost regardless of class frequency or BERT\textsubscript{EM} has a class preference that is only discoverable with macro-$F_1$.

We demonstrate that evaluation on micro-$F_1$ does not give adequate information about model performance on long-tail classes.
In Tables \ref{tab:tacred_results_table} and \ref{tab:semeval_results_table} we see that the model which performs better under micro-$F_1$ can either be significantly better or worse for classes with few samples.
The weighted-$F_1$ produces similar results to micro-$F_1$ except for RECENT.
Macro-$F_1$ on the other hand is very sensitive to model performance on single samples, e.g.\ the \emph{Entity-Destination(e2,e1)} class in SemEval.

The scores of our proposed schemes are in between the existing measures and might be the best indicators for robust generalization performance.
For all experiments, they produce similar results to each other.
This could just be a coincidence of the datasets, and is also indicated by Figure \ref{fig:tacred_weights}.
Overall, it might be fair to say that one of the former and latter measures is enough.
It would mean one measure that does weigh proportional to sample count (micro- or weighted-$F_1$), an intermediary measure (dodrans-$F_1$ or entropy-$F_1$) and macro-$F_1$.

PTR performs better for macro-$F_1$ on both datasets.
Its scores decrease less when classes are weighted more equally.
This suggests that it is a better model for classes with low sample counts.
\citet{lescao2021many} show that prompts can be worth hundreds of data points which would explain why the macro- and micro-$F_1$ scores are much closer together than for RECENT and BERT\textsubscript{EM}.

\section{Related Work}
\citet{chauhan2019reflex} do a thorough evaluation of their model and notice the significantly different performance measured by \emph{micro} and \emph{macro} statistics due to the class imbalance, suggesting that the choice of evaluation measure is crucial.
\citet{huang2020deepembeddings} further use the closeness between micro- and macro-$F_1$ scores to claim the stable performance of their model.

\citet{mille2021automatic} point out that evaluating with a single score favors overfitting.
They show different evaluation suites that can be created for a dataset.
\citet{bragg2021flex} address the disjoint evaluation settings across recent research threads in (few-shot) NLP and propose a unified evaluation benchmark which regulates dataset, sample size etc., but fail to take the evaluation measure into consideration, reporting only mean accuracy instead.
\citet{post2018call} criticises the inconsistency and under-specification in reporting scores.
This problem is also prevalent in RC where the $F_1$ weighting scheme is often not specified.

\citet{zhang2020demographics} show that bias from corpora persists for fine-tuned pre-trained language models.
These models struggle with rare phenomena.
For better performance debiasing with weighting is performed.
\citet{sogaard2021we} argue against using random splits.
They show that evaluating models with random splits is not a realistic setting but makes tasks easier by fixing the test data distribution to the train data distribution.

Long-tail evaluation is becoming more prominent in NLP research.
Models in deep learning tend to show a gap in performance between frequent and infrequent phenomena~\cite{rogers2021changing}.
Models in NLP have been shown to perform badly on specific subsets of data~\cite{zhang2020demographics}.

\citet{sokolova2009systematic} analyze measures for multi-class classification and present invariances regarding the confusion matrix.
\citet{gosgens2021good} also determine which class measures (including $F_1$) fulfil specific assumptions.
Further evaluation can be based on this.
Our weighting schemes for $F_1$ can be transferred to other measures that calculate a score for each class.

\section{Outlook}
We suggest creating and using a bidimensional leaderboard like~\citet{kasai2021bidimensional} where measures and models can be contributed.
To this end, benchmarking of RC models could be done on a centralized site where a model or test set predictions are submitted and measures are calculated automatically through a script.
For measures that modify weighting of classes and intra-class scoring, this does not require additional training computation.

Due to the reproducibility crisis~\cite{baker2016reproducibility}, not all state-of-the-art scores can be replicated.
Possible future work includes a comprehensive evaluation study of papers on leaderboards of RC tasks.
This would enable an in-depth discussion of strength and weaknesses (including reproducibility) of these models.

The analysis we present can also be extended to other NLP tasks with imbalanced datasets, such as named entity recognition~\cite{tjong2003introduction}, part-of-speech tagging~\cite{pradhan2013towards} and coreference resolution~\cite{pradhan2012conll}.

\section{Conclusion}
We criticise the current practice of reporting a single score when evaluating imbalanced RC datasets.
We propose a new framework to weight scores for multi-class evaluation of imbalanced datasets.
We provide two new weighting schemes, \emph{dodrans} and \emph{entropy}, which are positioned between \emph{class-weighted} and \emph{macro}.
In our experiments, we show that model performance on both TACRED and SemEval, especially on the long-tail relations, is not adequately captured by a single score.
Thus, we advocate the use of multiple weighing schemes when reporting model performance on imbalanced datasets.

\section*{Acknowledgments}
We would like to thank Nils Feldhus, Sebastian Möller, Lisa Raithel, Robert Schwarzenberg and the anonymous reviewers for their feedback on the paper. 
This work was partially supported by the German Federal Ministry of Education and Research as part of the project CORA4NLP (01IW20010) and by the German Federal Ministry for Economic Affairs and Climate Action as part of the project PLASS (01MD19003E).
Christoph Alt is supported by the Deutsche Forschungsgemeinschaft (DFG, German Research Foundation) under Germany's Excellence Strategy – EXC 2002/1 "Science of Intelligence" – project number 390523135.
\bibliography{custom}
\bibliographystyle{acl_natbib}
\appendix

\section{Implementation Details}
\label{sec:appendix-implementation}
To evaluate RECENT and PTR, we use the official code at \url{https://github.com//Saintfe/RECENT} (last updated on 01.10.2021) and \url{https://github.com/thunlp/PTR} (last updated on 20.11.2021).
Since the official code of BERT\textsubscript{EM} is not available, we implement this method using the HuggingFace Transformers library~\cite{wolf2020transformers} and PyTorch~\cite{paszke2019pytorch}, and make our code base available at \url{https://github.com/dfki-nlp/mtb-bert-em}.
To make our results reproducible, we randomly generated seeds \{9, 148, 378, 459, 687\} and employed these for all models in their 5 runs.

\section{Training Details}
\label{sec:appendix-training}
\subsection{RECENT}
We consider GCN as the base model.
Following the paper and the official code, we set the batch size to be $50$, the optimizer to be SGD with learning rate $0.3$, and the number of epochs to be $100$.
It takes a single RTX-A6000 GPU approximately 10 hours to complete all 5 runs on TACRED.

\subsection{BERT\textsubscript{EM}}
We use the pre-trained language model (PLM) \texttt{bert-large-uncased} from the HuggingFace model hub and directly fine-tune the model for the RC task, without matching-the-blank pre-training.
As the paper suggests, we set the batch size to be $64$, the optimizer to be Adam with learning rate $3\cdot 10^{-5}$, and the number of epochs to be $5$.
Additionally, we use the max sequence length of $512$. %
It takes a single RTX-A6000 GPU 30 minutes to complete all 5 runs on SemEval.

\subsection{PTR}
According to the paper and the official code base, we apply the same settings to evaluate both TACRED and SemEval:
We use the PLM \texttt{roberta-large} and set the max sequence length to be $512$, the batch size to be $64$, the optimizer to be Adam with learning rate $3\cdot 10^{-5}$, the weight decay to be $10^{-2}$, and the number of epochs to be $5$.
It takes 4 Quadro-P5000 GPUs 84 hours to complete 5 runs on TACRED, and it takes 8 Titan-V GPUs 9 hours on SemEval.

\end{document}